\documentclass[10pt,twocolumn,letterpaper]{article}

\usepackage{cvpr}
\usepackage{times}
\usepackage{epsfig}
\usepackage{graphicx}
\usepackage{amsmath}
\usepackage{amssymb}
\usepackage{float}
\usepackage{hyperref}
\begin{document}

%%%%%%%%% TITLE
\title{ SDeMorph: Towards Better Facial De-morphing from Single Morph}

\author{Nitish Shukla\\
Independent Researcher\\
{\tt\small nitishshukla86@gmail.com}
% \and
}

\maketitle
\thispagestyle{empty}

%%%%%%%%% ABSTRACT
\begin{abstract}
Face Recognition Systems (FRS) are vulnerable to morph attacks. A face morph is created by combining multiple identities with the intention to fool FRS and making it match the morph with multiple identities. Current Morph Attack Detection (MAD) can detect the morph but are unable to recover the identities used to create the morph with satisfactory outcomes. Existing work in de-morphing is mostly reference-based, i.e. they require the availability of one identity to recover the other. Sudipta et al. \cite{ref9} proposed a reference-free de-morphing technique but the visual realism of outputs produced were feeble. In this work, we propose SDeMorph (Stably Diffused De-morpher), a novel de-morphing method that is reference-free and recovers the identities of bona fides. Our method produces feature-rich outputs that are of significantly high quality in terms of definition and facial fidelity. Our method utilizes Denoising Diffusion Probabilistic Models
(DDPM) by destroying the input morphed signal and then reconstructing it back using a branched-UNet. Experiments on ASML, FRLL-FaceMorph, FRLL-MorDIFF, and SMDD datasets support the effectiveness of the proposed method.
\end{abstract}
\section{Introduction}

Face Recognition Systems (FRS) are widely deployed for person identification and verification for many secure access control applications. Amongst many, applications like the border control process where the face characteristics of a traveler are compared to a reference in a passport or visa database in order to verify identity require FRS systems to be robust and reliable. As with all applications, FRS is also prone to various attacks such as presentation attacks \cite{ref7}, electronic display attacks, print attacks, replay attacks, and 3D face mask attacks\cite{ref2,ref3,ref4,ref5,ref6}. Besides these, morphing attacks have also emerged as severe threats undermining the capabilities of FRS systems\cite{ref1,ref21}. In this paper, we focus on morph attacks.

Morph attack refers to generating a composite image that resembles closely to the identities it is created from. The morphed image preserves the biometric features of all participating identities \cite{ref22,ref23}. Morph attacks allow multiple identities to gain access using a single document\cite{ref24,ref25} as they can go undetected through manual inspection and are capable to confound automated FRS. In the recent past, deep learning techniques have been applied successfully to generate morphs. In particular, Generative Adversarial Networks (GAN) have shown tremendous success\cite{ref26,ref27,ref28,ref29,ref30}. Most of the morph generation techniques rely on facial landmarks where morphs are created by combining faces based on their corresponding landmarks\cite{ref31,ref32,ref33}. Deep learning methods simply eliminate the need for landmarks.

Morph Attack Detection (MAD) is crucial for the integrity and reliability of FRS. Broadly, MAD can be either a reference-free single-image technique\cite{ref34,ref35,ref36} or a reference-based differential-image technique\cite{ref37,ref38,ref39}. Reference-free methods utilize the facial features obtained from the input to detect whether the input is morphed or not whereas reference-based techniques compare the input image to a reference image which is typically a trusted live capture of the individual taken under a trusted acquisition scenario. 

MAD is essential from the security point of view but it does not reveal any information about the identities of the individuals involved in the making of morph. From a forensics standpoint, determining the identity of the persons participating in morph creation is essential and can help with legal proceedings. Limited work exists on face de-morphing and the majority of them are reference-based. In this paper, our objective is to decompose a single morphed image into the participating face images, without requiring any prior information on the morphing technique or the identities involved. We also make no assumption on the necessity of the input being a morphed image. Our work builds upon \cite{ref9} and aims to improve the results both visually and quantitatively. Overall, our contributions are as follows:
\begin{itemize}
    \item We propose SDeMorph to extract face images from a morphed input without any assumptions on the prior information. To the best of our knowledge, this is the first attempt to exploit DDPMs for facial image restoration in face morphing detection.
    \item A symmetric branched network architecture, that shares the latent code between its outputs is designed to de-morph the identity features of the bona fide participants hidden in the morphed facial image.
    \item We experimentally establish the efficacy of our method through extensive testing on various datasets. Results clearly show the effectiveness in terms of reconstruction quality and restoration accuracy.
\end{itemize}
The rest of the paper is organized as follows: Section \ref{background} gives a brief background on the diffusion process and formulates Face de-morphing. Section \ref{method} introduces the rationale and proposed method. Section \ref{result} outlines the implementation details, experiments, and results. Finally, Section \ref{summary} concludes the paper.
\begin{figure}
    \centering
    \includegraphics[width=\columnwidth]{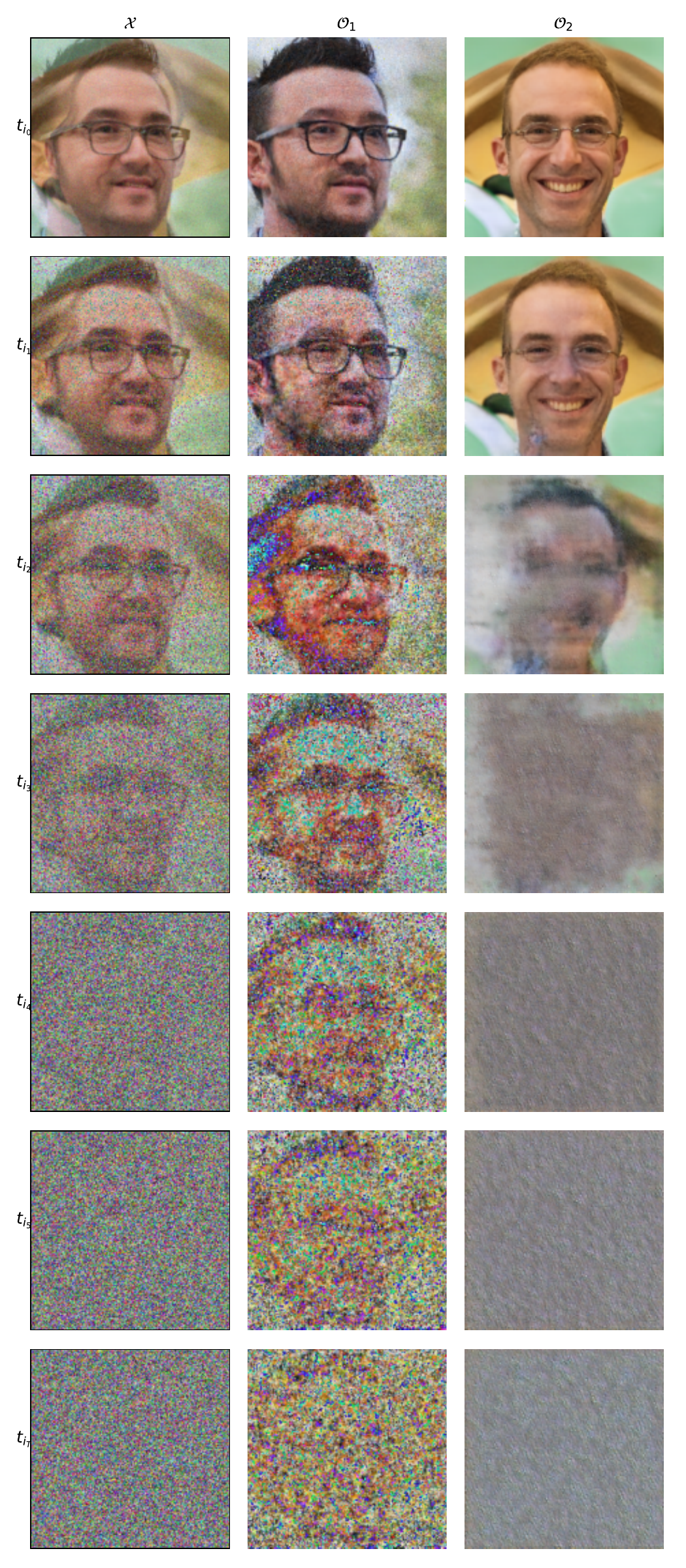}
    \caption{Illustration of the noise schedule and reconstruction performed by the proposed method. The scheduler adds noise to input (Left column) until the input signal is destroyed. (Middle, Right column) The model aims to predict the noise schedule to reconstruct the bona fides. The final outputs are extracted at $t=0$.}
    \label{fig:diffusion}
\end{figure}
\section{Background}
\label{background}
\subsection{Denoising Diffusion Probabilistic Models (DDPM)}
On a high level, DDPMs\cite{ref8} are latent generative models that learn to produce or recreate a fixed Markov chain $x_1,x_2,...,x_T$. The forward Markov transition, given the initial data distribution $x_0\sim q(x_0)$, adds gradual Gaussian noise to the data according to a variance schedule $\beta_1,\beta_2,....,\beta_T$, that is,
\begin{center}
   \begin{equation}
    % q(x_{1:T}|x_0)=\prod_{i=1}^T q()
    q(x_t|x_{t-1})=\mathcal{N}(x_t;\sqrt{1-\beta}x_{t-1},\beta_t\mathbb{I})
\end{equation} 
\end{center}

The conditional probability $q(x_t|x_0)$ (diffusion) and $q(x_{t-1}|x_0)$ (sampling), can be expressed using Bayes’ rule and Markov property as 
    \begin{equation}
        q(x_t|x_0)=\mathcal{N}(x_t;\sqrt{\Bar{\alpha}}x_0;(1-\Bar{\alpha})\mathbb{I}) ,\ \ t=1,2,...,T
    \end{equation}
    \begin{equation}
        q(x_{t-1}|x_t)=\mathcal{N}(x_{t-1};\Bar{\mu}(x_t,x_0);\tilde{\beta}\mathbb{I}) ,t=1,2,...,T
    \end{equation}

where $\alpha_t=1-\beta_t$, $\Bar{\alpha}_t=\prod_{s=1}^t\alpha_s$, $\tilde{\beta}=\frac{1-\Bar{\alpha}_{t-1}}{1-\Bar{\alpha}_t}\beta_t$ and 

$\tilde{\mu}(x_t,x_0)=\frac{\sqrt{\Bar{\alpha}_t}\beta_t}{1-\Bar{\alpha}_t}x_0 + \frac{\sqrt{\alpha_t}(1-\Bar{\alpha}_{t-1})}{1-\Bar{\alpha}_t}x_t $. 

DDPMs generate the Markov chain by using the reverse process having prior distribution as $p(x_T)=\mathcal{N}(x_T;0,\mathbb{I})$ and Gaussian transition distribution as
\begin{equation}
    \label{eq:rev}
    p_\theta(x_{t-1}|x_t)=\mathcal{N}(x_{t-1};\mu_\theta(x_t,t),\sum_{\theta}(x_t,t)) , \ \ t=T,..,1
\end{equation}
The parameters $\theta$ are learned to make sure that the generated reverse process closely mimics the noise added during the forward process.
The training aims to optimize the objective function which has a closed form given as
the KL divergence between Gaussian distributions. The objective can be simplified as $\mathbb{E}_{x,\epsilon \sim \mathcal{N}(0,1),t }  [ || \epsilon -\epsilon_\theta(x_t,t) ||_2^2]$.
\subsection{Face Morphing}
Face morphing refers to the process of combining two faces, denoted as $\mathcal{I}_1$ and $\mathcal{I}_2$ to produce a morphed image, $\mathcal{X}$ by aligning the geometric landmarks as well as blending the pixel level attributes. The morphing operator $\mathcal{M}$ defined as
\begin{equation}
    \mathcal{X}=\mathcal{M}(\mathcal{I}_1,\mathcal{I}_2)
\end{equation}
aims to produce $\mathcal{X}$, such that the biometric similarity of the morph and the bona fides is high i.e. for the output to be called a successful morph attack, $\mathcal{B}(\mathcal{X},\mathcal{I}_1)>\tau$ and $\mathcal{B}(\mathcal{X},\mathcal{I}_2)>\tau$ should hold, where $\mathcal{B}$ is a biometric comparator and $\tau$ is the threshold value.

Initial work on de-morphing\cite{ref39} used the reference of one identity to recover the identity of the second image. The authors also assumed prior knowledge about landmark points used in morphing and the  parameters of the morphing process. FD-GAN\cite{ref19} also uses a reference to recover identities as in the previous method. It uses a dual architecture and attempts to recover the first image from the morphed input using the second identity's image. It then tries to recover the second identity using the output of the first identity by the network.
This is done to validate the effectiveness of their generative model.

\begin{figure*}
    \centering
    \includegraphics[scale=0.35]{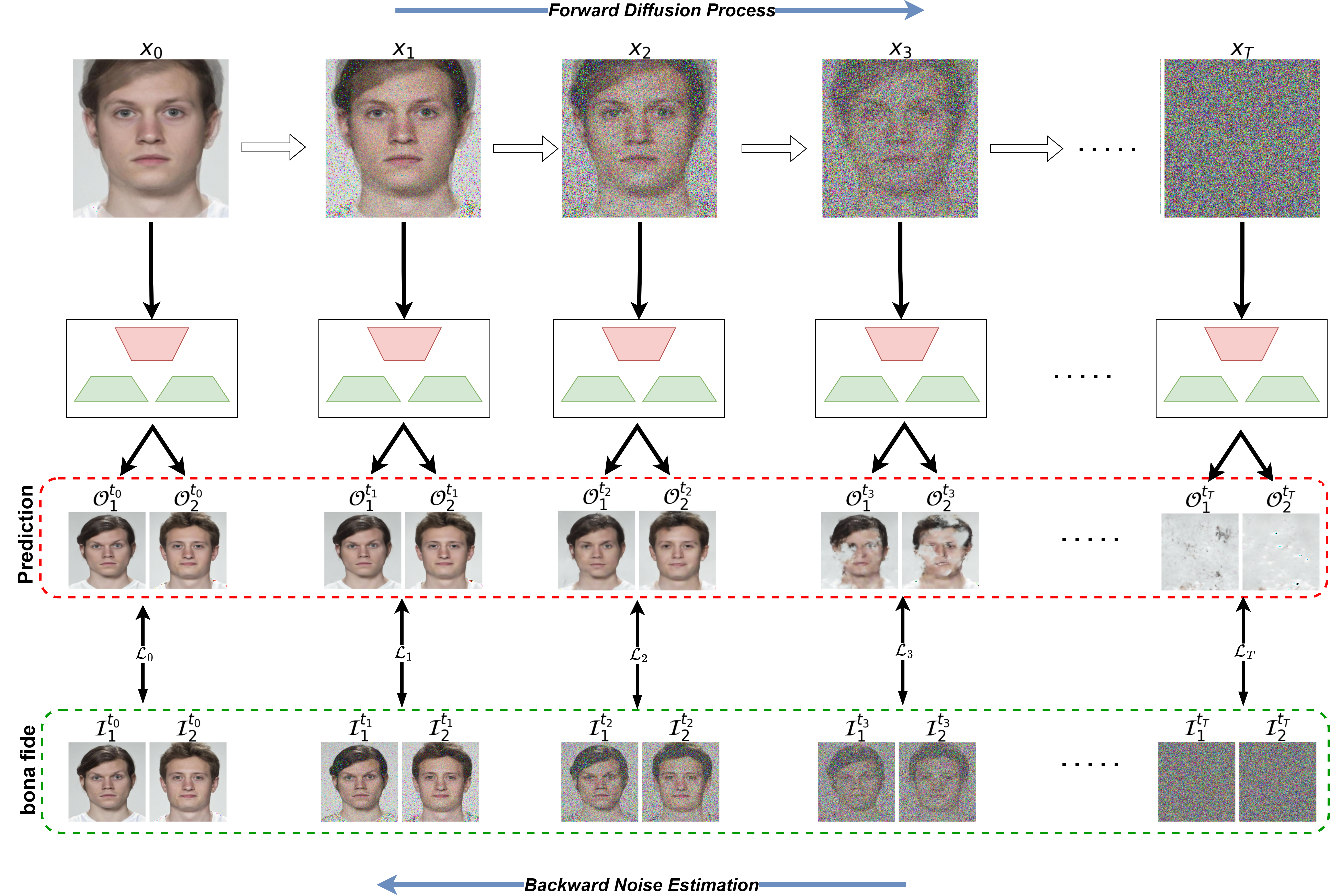}
    \caption{Illustration of the architecture of the proposed method. A diffusion process incrementally adds noise to the input. A branched-UNet is used to estimate the noise schedule.}
    \label{fig:architecture}
\end{figure*}

\section{Methodology}
\label{method}
\subsection{Rationale}
In \cite{ref9}, the authors decompose the morphed image into output images using a GAN that is composed of a generator, a decomposition critic, and two markovian discriminators. Inspired by the work, we propose a novel method that takes a morphed image $\mathcal{X}$ and decomposes it into output images $\mathcal{O}_1$ and $\mathcal{O}_2$. The goal of the method is to produce outputs similar to the bona fides(BF), $\mathcal{I}_1$ and $\mathcal{I}_2$. The method also works with non-morphed images, i.e. if the input is a non-morphed image, the method would generate outputs very similar to the inputs ($\mathcal{O}_1\approx \mathcal{O}_2  \approx \mathcal{X}$). The task of decomposition of morphed images can be well equated with the problem of separating two signals which has been studied extensively. Among many, we can mention independent component analysis (ICA) \cite{ref10,ref11}, morphological component analysis (MCA) \cite{ref14,ref15,ref16}, and robust principal component analysis \cite{ref12,ref13}. These methods rely on strong prior assumptions such as independence, low rankness, sparsity, etc. However, the application of these techniques in de-morphing faces is difficult because such strong prior assumptions are typically not met. Motivated by the above-mentioned issues, we propose a novel method that is reference-free, i.e. takes a single morphed image and recovers the bona fides images used in creating the morph. In this paper, We closely follow the methodology in \cite{ref8} with two changes 1) a \textit{branched}-UNet is used instead of regular UNet and 2) cross-road loss is implemented. We explain both in \ref{proposed}.

\subsection{Proposed Method}
\label{proposed}
The morphing operator $\mathcal{M}(\cdot,\cdot)$, typically involves highly intricate and non-linear warping image editing functions which make de-morphing from a single image an ill-posed problem. We adopt a generative diffusion probabilistic model that iteratively adds noise to the input until the input signal is destroyed. The reverse process learns to recover the input signal from the noise.
\subsubsection{Forward diffusion Process}
The forward diffusion process consists of adding a small amount of Gaussian noise to the input in steps ranging from $0$ to $T$. This result in the input sequence $x_0,x_1,x_2,....,x_T$, where $x_0$ is the unadulterated sample. As $T\to\infty$, $x_T$ becomes equivalent to an isotropic Gaussian distribution. The step size is controlled by the variance schedule $\{ \beta_t \in(0,1) \}_{t=0}^T$. The forward process is typically fixed and predefined.
During the forward process, we add the aforementioned noise schedule to the morphed image until the signal degenerated into pure noise as illustrated in Figure \ref{fig:diffusion} (first column).
\subsubsection{Reverse sampling process}
The goal of the learning is to estimate the noise schedule, i.e. the amount of noise added at time step $t$. We follow a similar setup as in \cite{ref8}. A deep learning network is used to realize $p_\theta$ in Equation \ref{eq:rev}. In this paper, we have employed a \textit{branched}-UNet which is trained to predict the parameters used during the forward process. A branched-UNet shares the same latent code with both of its outputs. This enables the model to output images that are semantically closer to the input. Figure \ref{fig:diffusion} illustrates this, at time $t$, the UNet takes input, the noisy image, and tries to reconstruct the noisy version of the ground truth (second and third column). Finally, the clean output is extracted at $t=0$.

\subsubsection{Loss function}
The sampling function $f$ used to estimate the reverse process is trained with the ``cross-road'' loss defined as
\begin{equation}
\begin{aligned}
\mathcal{L} =\sum_{t}\min &[\mathcal{L}_1^t(\mathcal{I}_1^t,\mathcal{O}_1^t)+ \mathcal{L}_1^t(\mathcal{I}_2^t,\mathcal{O}_2^t),\\
      &     \mathcal{L}_1^t(\mathcal{I}_1^t,\mathcal{O}_2^t)+ \mathcal{L}_1^t(\mathcal{I}_2^t,\mathcal{O}_1^t)
    ]
\end{aligned}
\end{equation}
where $\mathcal{L}_1$ is the per-pixel loss. $\mathcal{I}_i^t, \mathcal{O}_i^t$, $i=1,2$ are the noisy inputs and outputs of the sampling process respectively at time step $t$. The reason for using cross-road loss is that the outputs of the reverse process lack any order. Having that, it is not guaranteed that $\mathcal{O}_1$ corresponds to $\mathcal{I}_1$ and $\mathcal{O}_2$ to $\mathcal{I}_2$. Therefore, we consider only 2 possible scenarios and incorporate that into the loss. Taking the minimum of both cases ensures that the correct pairing is done. The loss encourages the sampling process to estimate the noise added during the forward process by forcing the outputs to be visually similar to noisy inputs at time $t$. 
\begin{figure*}[h!]
     \centering
     \begin{subfigure}[b]{0.45\textwidth}
         \centering
         \includegraphics[width=\textwidth]{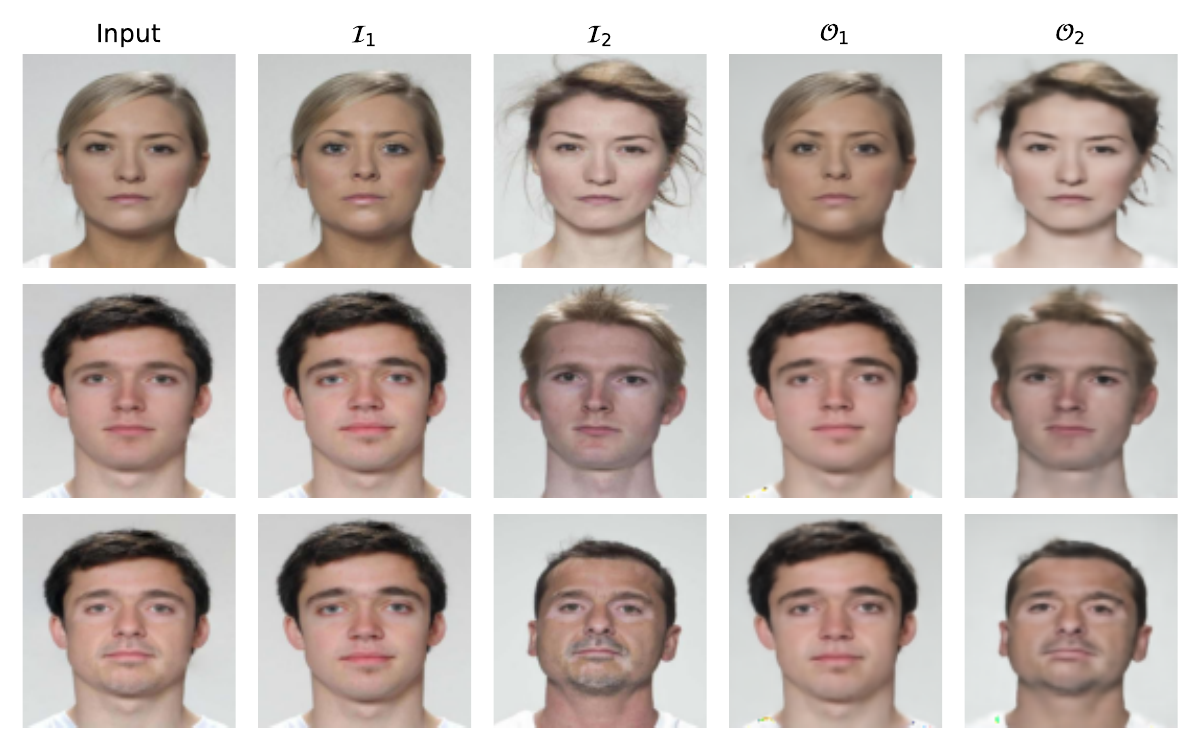}
         % \label{fig:y equals x}
     \end{subfigure}
     \begin{subfigure}[b]{0.45\textwidth}
         \centering
         \includegraphics[width=\textwidth]{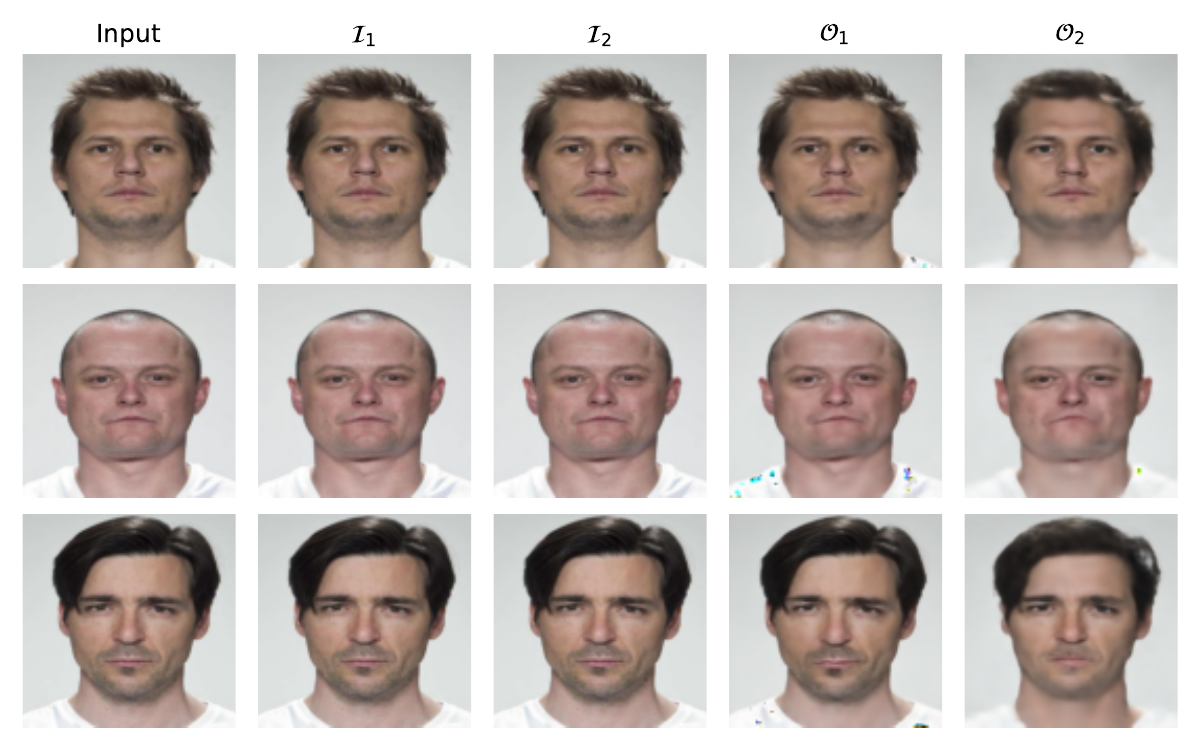}
         % \caption{$y=3\sin x$}
         % \label{fig:three sin x}
     \end{subfigure}
     
     \caption{Illustration of the de-morphed images produced using the proposed method on AMSL dataset. Here, ``input" refers to the morphed image obtained from $\mathcal{I}_1$ and $\mathcal{I}_2$. The outputs are denoted as $\mathcal{O}_1$ and $\mathcal{O}_2$. (Left) The generated images achieve remarkably high similarity and visual realism when the input is indeed a morphed image. (Right) The model replicates the input when given an un-morphed image, i.e. $\mathcal{X}=\mathcal{I}_1=\mathcal{I}_2$ }
     \label{fig:asml_sim}
    \end{figure*}
\begin{figure*}
     \centering
     \begin{subfigure}[b]{0.45\textwidth}
         \centering
         \includegraphics[width=\textwidth]{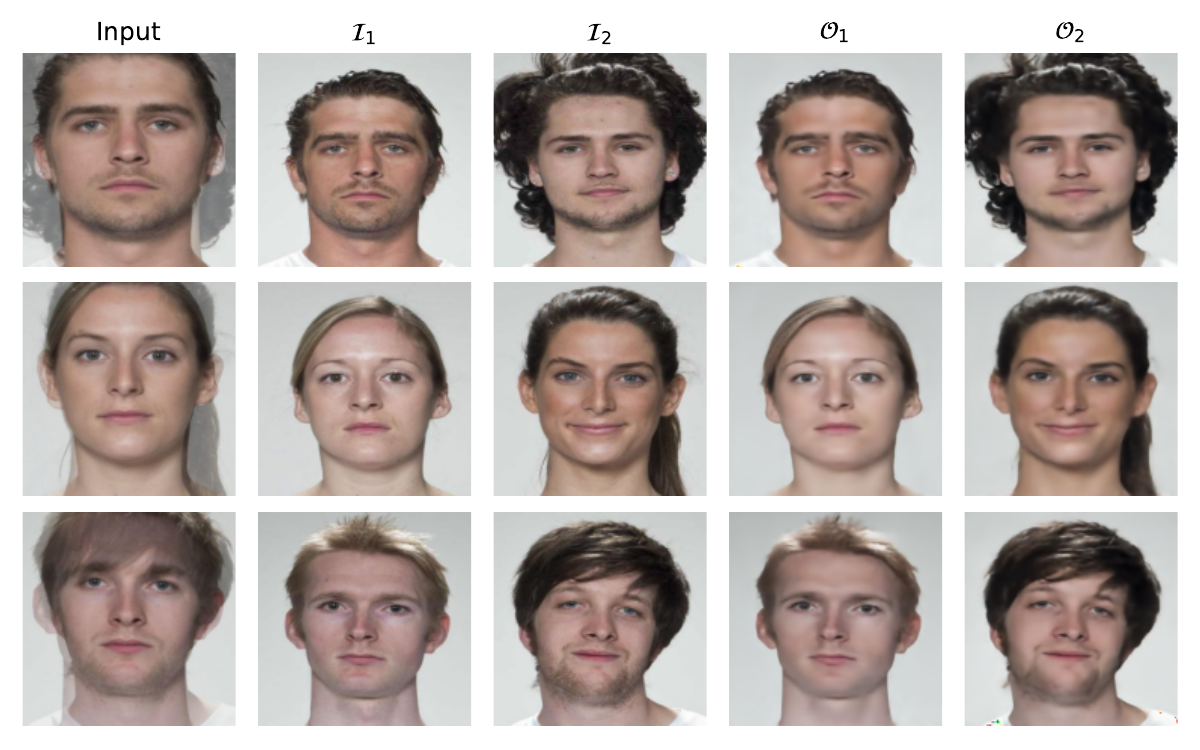}
         % \caption{$y=x$}
         \label{fig:y equals x}
     \end{subfigure}
     \hfill
     \begin{subfigure}[b]{0.45\textwidth}
         \centering
         \includegraphics[width=\textwidth]{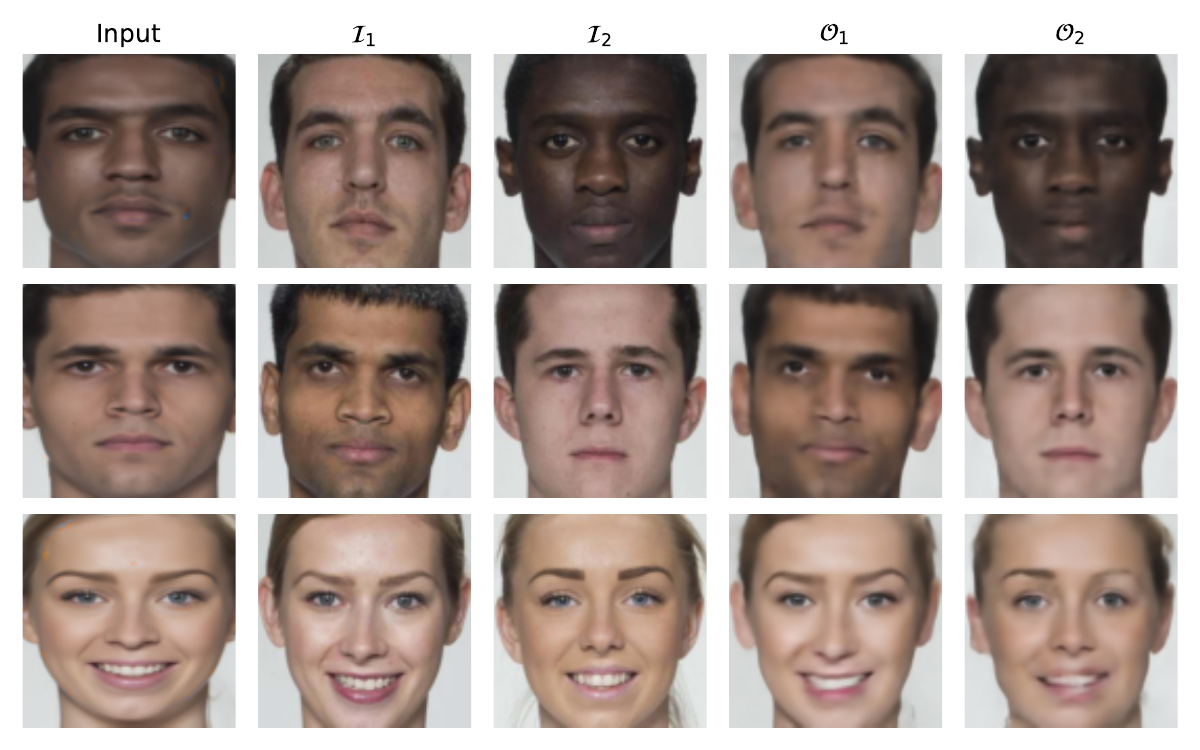}
         % \caption{$y=3\sin x$}
         \label{fig:three sin x}
     \end{subfigure}
     \hfill
     \begin{subfigure}[b]{0.45\textwidth}
         \centering
         \includegraphics[width=\textwidth]{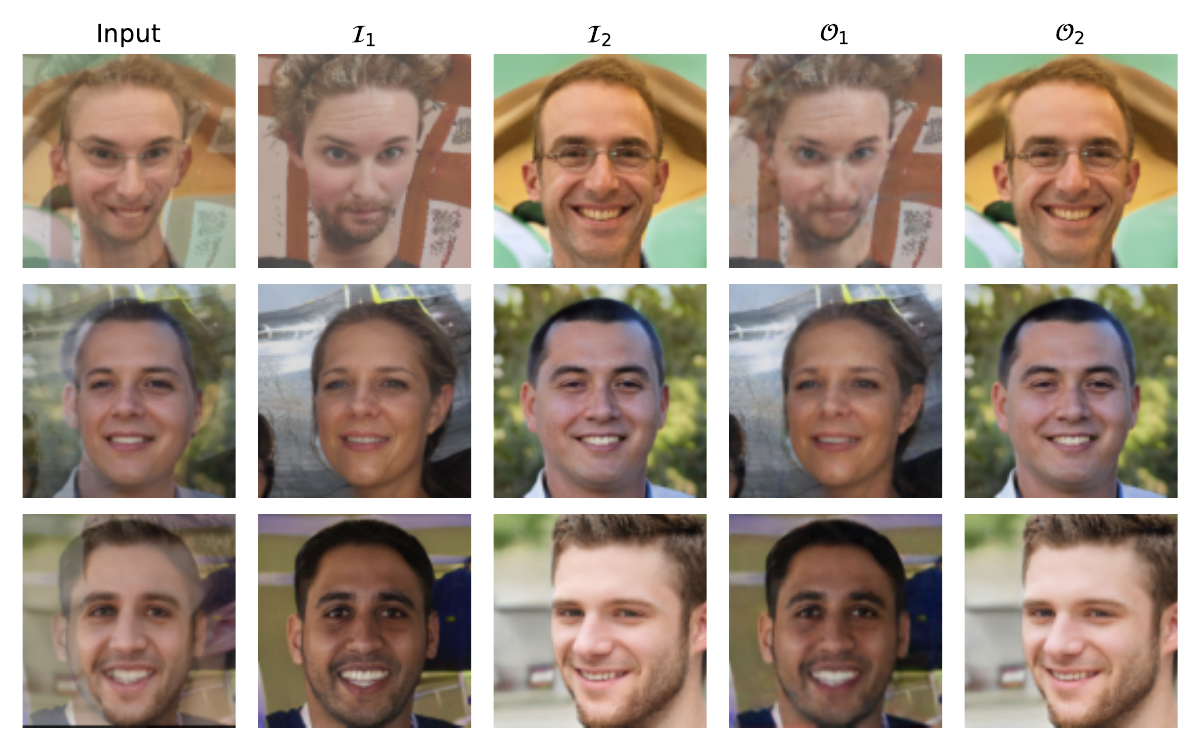}
         % \caption{$y=5/x$}
         \label{fig:five over x}
     \end{subfigure}
        \caption{Faces reconstructed by the proposed method on (Top Left) FRLL-Facemorph, MorDIFF (Top Right) and SMDD datasets (Bottom). The reconstructed images have very high similarity with their corresponding BF samples.}
        \label{fig:three graphs}
\end{figure*}
%%%%%%%%% BODY TEXT
\section{Experiments and Results}
\label{result}
\subsection{Datasets and Preprocessing}
We perform our experiment with 4 different morphing techniques on the following datasets:

\noindent \textbf{AMSL face morph dataset} The dataset contains 2,175 morphed images belonging to 102 subjects captured with neutral and smiling expressions. Not all of the images are used to create morphed images. We randomly sample $80\%$  of the data as our training set and the remaining is used as test set. This setting is maintained throughout the datasets used in this paper.

\noindent \textbf{FRLL-Morphs} The dataset is constructed using the Face Research London Lab dataset. The morphs are created using 5 different morphing techniques, namely, OpenCV (OCV), FaceMorpher (FM), Style-GAN 2 (SG), and WebMorpher (WM). We conduct our experiments on FaceMorpher morphs. Each morph method contains 1222 morphed
faces generated from 204 bona fide samples. All the images are generated using only frontal face images.

\noindent \textbf{MorDIFF} The MorDIFF dataset is an extension of the SYN-MAD 2022 dataset using the same morphing pairs. Both SYN-MAD and MorDIFF are based on FRLL dataset. The dataset contains 1000 attack images generated from 250 BF samples, each categorized on the basis of gender(male/female) and expression(neutral/smiling).

\noindent\textbf{SMDD} The dataset consists of 25k morphed images and 15k BF images constructed from 500k synthetic images generated StyleGAN2-ADA trained on Flickr-Faces-HQ Dataset (FFHQ) dataset. The evaluation dataset also has an equal number of morphed and bona fide images.

In this paper, we have used morphed images for training but testing is done on both morphed and unmorphed images.

\begin{figure*}
    \centering
    \includegraphics[scale=0.46]{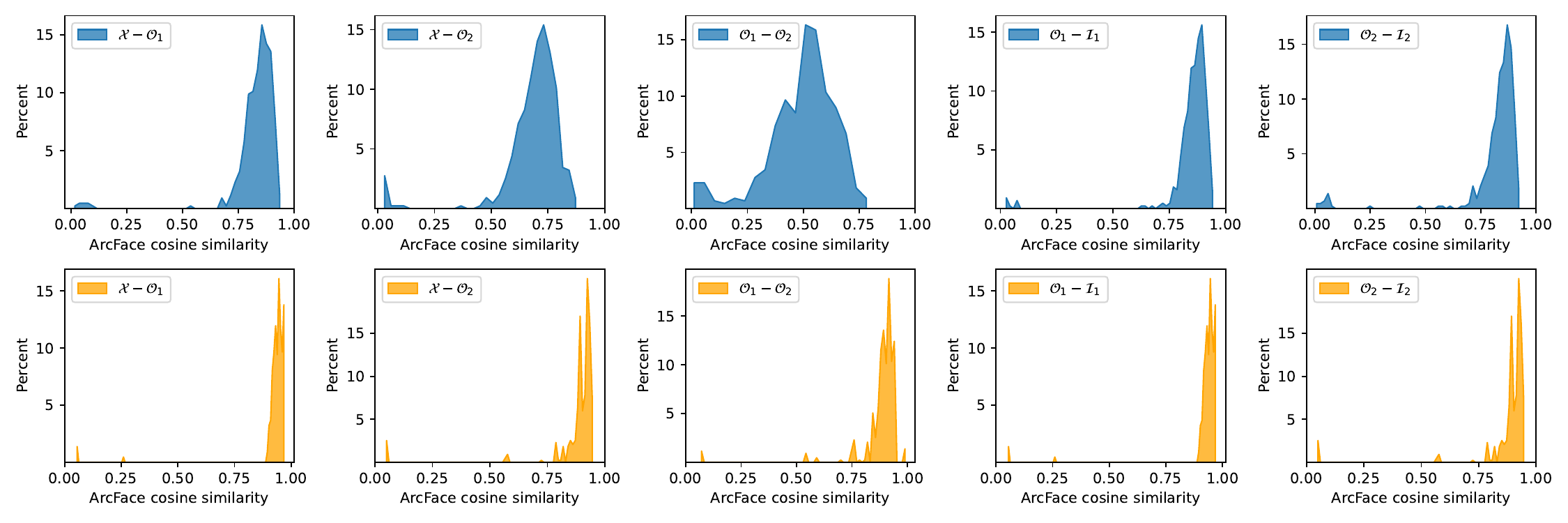}
    \caption{Distribution of ArcFace cosine similarity on AMSL dataset. (Top Row) The proposed method achieves high facial similarity on the input-output pair $(\mathcal{I}_i,\mathcal{O}_i), i=1,2$ (Column 4,5) in the morphed case. (Bottom Row) A similar trend is observed in the unmorphed case when $\mathcal{X}=\mathcal{I}_1=\mathcal{I}_2$.  }
    \label{fig:amsl_arcface}
\end{figure*}
% \begin{table*}[]
%     \centering
%     \begin{tabular}{cc}
%         \includegraphics[scale=0.4]{images/amsl_morphed.pdf}  &\includegraphics[scale=0.4]{images/amsl_unmorphed.pdf}  \\
%     % \label{asml_sim}
%     \end{tabular}
%     \caption{illustration of the de-morphed images produced using the proposed method on AMSL dataset. Here, ``input" refers to the morphed image obtained from $\mathcal{I}_1$ and $\mathcal{I}_2$. (Left) The generated images achieve remarkably high similarity and visual realism when the input is a morphed image. (Right) The model replicates the input when given an un-morphed image, i.e. $\mathcal{X}=\mathcal{I}_1=\mathcal{I}_2$  }
%     \label{asml_sim}
% \end{table*}

\begin{figure*}[h!]
    \begin{subfigure}[b]{\textwidth}
        \includegraphics[width=\textwidth]{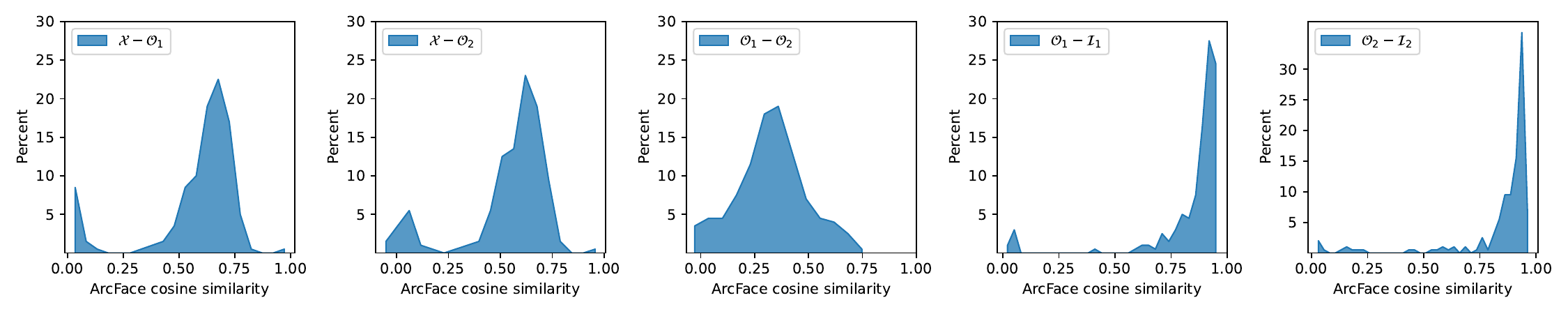}
    \end{subfigure}
        \begin{subfigure}[b]{\textwidth}
        \includegraphics[width=\textwidth]{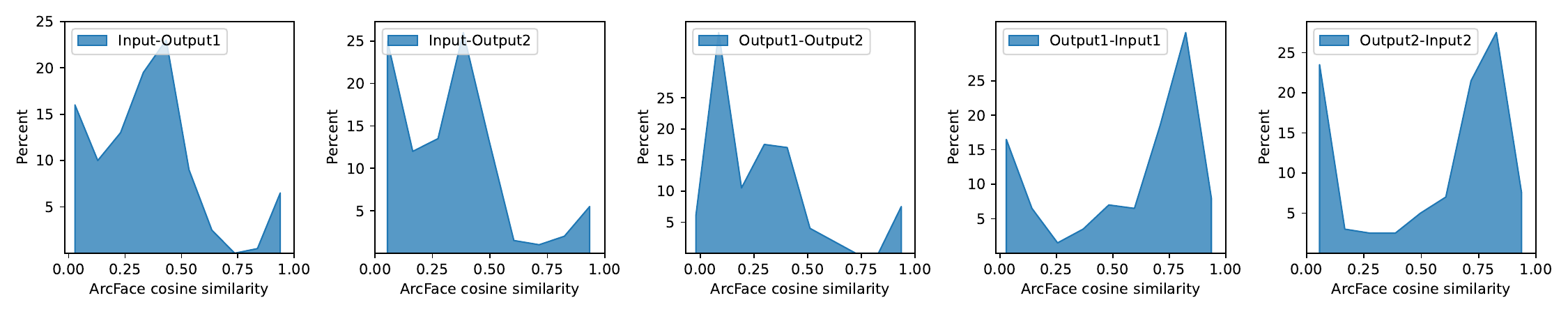}
    \end{subfigure}
    \begin{subfigure}[b]{\textwidth}
        \includegraphics[width=\textwidth]{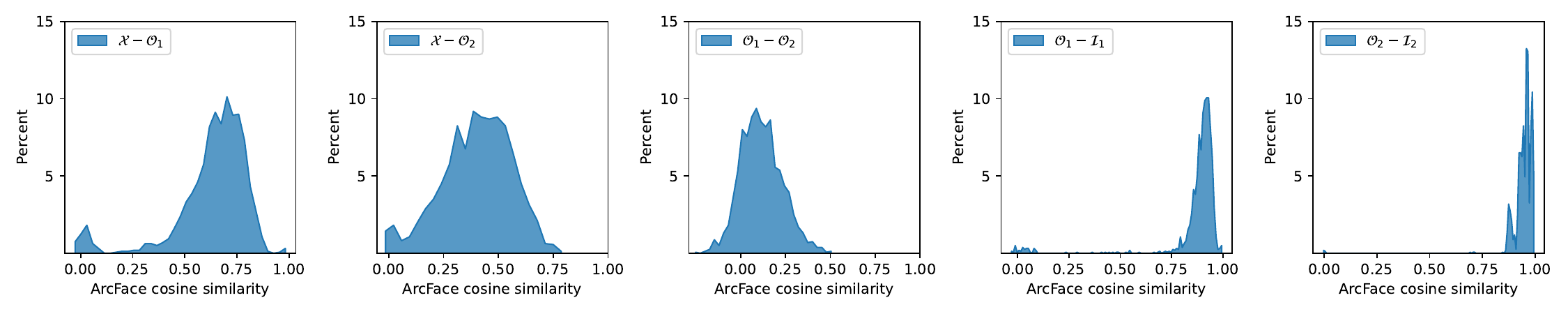}
    \end{subfigure}

    \caption{Distribution of ArcFace cosine similarity on FRLL-FaceMorph(Top), MorDIFF(Middle) and SMDD dataset(Bottom). In all three cases, the proposed method outputs high-quality images having near-perfect similarity to BF samples.}
    \label{fig:arcface_similarity}
\end{figure*}

\begin{figure}
    \centering
    \includegraphics[width=\columnwidth]{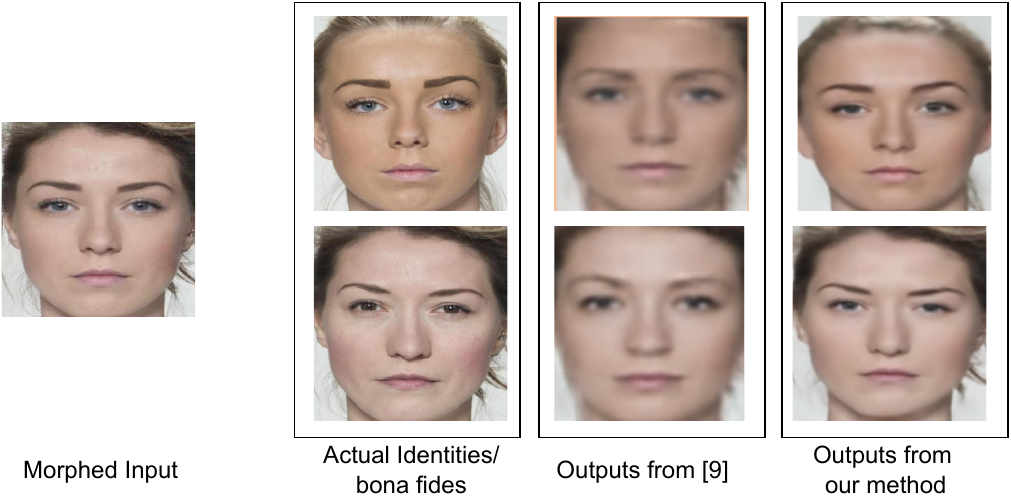}
    \caption{We compare the visual quality of outputs from our method to our closest competitor \cite{ref9}. The first column is the morphed input and the second column are the original identities used to create the morph. The remaining two columns are the outputs generated from \cite{ref9} and our method. In terms of facial features, our method generates face images that are much similar to ground truth. Moreover, the images generated are also of superior quality.  }
    \label{fig:compare}
\end{figure}
\subsection{Implementation Details}
Throughout our experiments, we set $T=400$ for MorDIFF and $T=300$ for remaining datasets. Our method does not generate data, thus a smaller value of $T$ is preferred. The beta schedule for variances in the forward process is scaled linearly from $\beta_0=10^{-4}$ to $\beta_T=0.02$. With the schedule, the forward process produces $\mathcal{X}^t, \mathcal{I}_1^t$ and $\mathcal{I}_2^t$, $t=0,1,..,T$, the noisy morphed image and corresponding bona fides respectively.

The reverse process is realized by a \textit{branched}-UNet. The UNet has an encoder consisting of convolution layers followed by batch normalization \cite{ref40}. To embed the time information, we use Transformer sinusoidal position embedding \cite{ref17}. The UNet contains 2 identical decoders consisting of transpose convolutions and batch norm layers. Both the decoders share the same latent space and identical skip connections from the encoder layer. The training was done using Adam optimization with an initial learning rate of $10^{-3}$ for $300$ epochs. 
To quantitatively compare the generated faces and ground truth, we use an open-source implementation of ArcFace\cite{ref18} network as a biometric comparator with cosine distance as the similarity measure.

\subsection{Results}
We evaluate our method on both morphed and non-morphed images. In ideal conditions, the method outputs bona fides when the input is indeed a morphed image and replicates the input twice when an unmorphed image is inputted. We evaluate our de-morphing method both quantitatively and qualitatively. We visualize the reconstructions of bona fides by the proposed method on AMSL dataset in Figure \ref{fig:asml_sim}. The first column ``input" is the morphed image, and the next two columns ($\mathcal{I}_1,\mathcal{I}_2$) are the bona fides used to construct $\mathcal{X}$. Finally, the remaining two columns ($\mathcal{O}_1,\mathcal{O}_2$) are the outputs produced by the method at $t=0$. We observe that our method produces realistic images that visually resemble the ground truth. The method not only learns the facial features but also features like hairstyle (first row) and skin features like vitiligo (last row). The produced images are also significantly sharper compared to existing methods as illustrated in figure \ref{fig:compare}.
The right set in Figure \ref{fig:asml_sim} are reconstruction on unmorphed images ($\mathcal{X}=\mathcal{I}_1=\mathcal{I}_2$). We observe that the method successfully replicates the input. Note that the outputs produced are not identical to each other but mere variations of the same unmorphed input. The method manages to replicate the unmorphed image with high facial fidelity despite having never been trained with it (the model is only trained on morphed images). We visualize similar results on FRLL-Morph, MorDIFF, and SMDD datasets in Figure \ref{fig:three graphs}. We observe similar visual results on these datasets with minor artifacts produced on MorDIFF. We believe that this is because of the usage of diffusion autoencoder to perform the MorDIFF attack which makes our method a direct inverse of the attack. 

\begin{table}[h]
    \centering
    \begin{tabular}{|p{1.5cm}|p{1.2cm}|p{1.2cm}|p{1.2cm}|p{1.2cm}|}
    \hline
        Restoration Accuracy & ASML &FRLL FaceMorph & FRLL MorDIFF &SMDD\\
        \hline
        Subject 1 &97.70\% &96.00\% & 78.00\% & 96.57\%\\
        Subject 2 & 97.24\%& 99.50\%& 74.00\% & 99.37\%\\
        \hline
    \end{tabular}
    
    \caption{We compute the restoration accuracy between bona fide and  generated samples ($\mathcal{O}_i,\mathcal{I}_i$), $i=1,2$.  }
    \label{tab:rest}
\end{table}

% \begin{table*}
%     \centering
%     \begin{tabular}{c}
%          \includegraphics[scale=0.4]{images/amsl_facemorpher_arcface.pdf}\\
%          \includegraphics[scale=0.4]{images/mordiff_arcface.pdf}\\
%          \includegraphics[scale=0.4]{images/smdd_arcface.pdf}
%     \end{tabular}
%     \caption{Distribution of ArcFace cosine similarity on FRLL-FaceMorph(Top), MorDIFF(Middle) and SMDD dataset(Bottom). In all three cases, the proposed method outputs high quality images having near perfect similarity to BF samples.}
%     \label{tab:arcface_similarity}
% \end{table*}

We also compare the generated faces using a biometric comparator $\mathcal{B}$ to validate that our method is not generating faces with arbitrary features (i.e. arbitrary faces). We employ ArcFace as comparator $\mathcal{B}$ and cosine distance as measure of similarity, large scores indicate higher facial similarity. We compute the ArcFace similarity between the following combinations between input and outputs: $(\mathcal{X},\mathcal{O}_1),(\mathcal{X},\mathcal{O}_2),(\mathcal{O}_1,\mathcal{O}_2),(\mathcal{O}_1,\mathcal{I}_1)$ and $(\mathcal{O}_2,\mathcal{I}_2)$. On AMSL dataset, Figure \ref{fig:amsl_arcface} visualizes the cosine similarity plots, the $X$ axis represents the cosine similarity score between the pair of images, and $Y$ axis is the percentage of test pairs attaining the similarity score. The top row represents the morphed case whereas the bottom row are plots pertaining to unmorphed images. We observe that the similarity plots of $(\mathcal{O}_i,\mathcal{I}_i), i=1,2$ are heavily skewed towards the similarity of $1$ (column 4,5). This indicates that the generated and corresponding bona fide sample belongs to the same person. Moreover, we observe that the distribution of ($\mathcal{O}_1,\mathcal{O}_2$) is centered around $0.5$, indicating that our model outputs images that are facially distinct within themselves. The bottom row of Figure \ref{fig:amsl_arcface} contains the cosine similarity plots on unmorphed images ($\mathcal{X}\approx\mathcal{I}_1\approx\mathcal{I}_2$) from AMSL dataset. In this case, we see that all the similarity plots are identical and skewed towards $1$. This indicates that the method replicates the input image as both its outputs. Similar plots on FRLL-FaceMorph, MorDIFF, and SMDD datasets are presented in Figure \ref{fig:arcface_similarity}. 

Finally, to quantitatively measure the efficacy of our method, we compute the restoration accuracy\cite{ref19} defined as the fraction of generated images that correctly match with their corresponding bona fide but does not match with the other bona fide (i.e. each output has exactly one matching bona fide) to the total number of test samples. 
% To further explain, among input and output pairs $(\mathcal{I}_1,\mathcal{I}_2)$ and $(\mathcal{O}_1,\mathcal{O}_2)$, we call it a match if the following holds
% \begin{equation}
%     [\mathcal{B}(\mathcal{I}_1,\mathcal{O}_1)>\tau] \land [\mathcal{B}(\mathcal{I}_1,\mathcal{O}_2)<\tau]
%     \land [\mathcal{B}(\mathcal{I}_1,\mathcal{O}_2)<\tau]
%     \land [\mathcal{B}(\mathcal{I}_1,\mathcal{O}_2)<\tau]
% \end{equation}

We use publicly available Face++\cite{ref20} API to compare the bona fides and generated faces. The restoration accuracy is reported in Table \ref{tab:rest}. (i) \textbf{ASML}: Our method achieves restoration accuracy of $97.70\%$ for Subject 1 and $97.24$ for Subject 2. This means that over $97\%$ of generated images correctly matched with their corresponding BF but didn't match with the other BF. (ii) \textbf{FRLL-FaceMorph}: $96.00\%$ for Subject 1 and $99.50\%$ for Subject 2. (iii) \textbf{FRLL MorDIFF}: $78.00\%$ for Subject 1 and $74.00\%$ for Subject 2 and finally (iv) \textbf{SMDD}: $96.57\%$ for Subject 1 and $99.37\%$ for Subject 2. The results indicate that our method performs well in terms of restoration accuracy.

\noindent\textbf{MAD Performance: } Apart from restoration accuracy, we also perform MAD experiments and measure the performance on the metric of APCER@5\%BPCER. We report the results in Table \ref{tab:rest}. We observe a value of 2.08\% for ASML dataset, 4.12\% for FaceMorph, 12.18\% for MorDiff and 6.41\% for SMDD dataset. Lower values indicate that our method separates the distribution of morphs and bona fides significantly.

% APCER, BPCER and ACER and report them in Table \ref{tab:rest}. Note that our method simply copies the input image as both of its output if the input is an unmorphed image. Using this fact, \textbf{we classify input image as a morph if the output faces produced by our method do not match}. We use \texttt{face recognition}\cite{fr} API to detect and compare faces. Furthermore, we compute ROC and DET scores and plot them in Figure \ref{fig:roc-det}.

\begin{table}[]
    \centering
    \begin{tabular}{|p{3cm}|c|}
    \hline
         Dataset& APCER @5\%BPCER \\
        \hline
        ASML &2.08  \\
        FRLL FaceMorph & 4.12\\
        FRLL MorDIFF &12.18 \\
        SMDD & 6.41 \\
        \hline
    \end{tabular}
    
    \caption{Morph Attack Detection accuracy. We compute APCER@5\%BPCER  }
    \label{tab:rest}
\end{table}

% \begin{figure}[h]
%     \centering
%     \includegraphics[scale=0.31]{images/det.pdf}
%     \caption{We  }
%     \label{fig:roc-det}
% \end{figure}

% {\small
% \bibliographystyle{ieee}
% % \bibliography{egbib}
% \bibitem[]{ref10}
% A. Hyvarinen, J. Karhunen, and E. Oja, Independent Component Anal- [53] ysis, vol. 46. New York, NY, USA: Wiley, 2004.
% }

\section{Summary}
\label{summary}
In this paper, we have proposed a novel de-morphing method to recover the identity of bona fides used to create the morph. Our method is reference-free, i.e. the method does not require any prior information on the morphing process which is typically a requirement for existing de-morphing techniques. We use DDPM to iteratively destroy the signal in the input morphed image and during reconstruction, learn the noise schedule for each of the participating bona fides. To train, we employ an intuitive ``cross-road" loss that automatically matches the outputs to the ground truth. We evaluate our method on AMSL, FRLL-FaceMorph, MorDIFF, and SMDD datasets resulting in visually compelling reconstructions and excellent biometric verification performance with original face images. We also show that our method outperforms its competitors in terms of quality of outputs (i.e. produces sharper, feature rich images) while keeping high restoration accuracy.  
\bibliographystyle{unsrt}  
\bibliography{references}

\end{document}